\documentclass{llncs}
\usepackage{makeidx}  
\usepackage{url}      
\usepackage{graphicx}
\usepackage{amsfonts}
\usepackage{amsmath}
\usepackage{algorithm}
\usepackage{algpseudocode}
\usepackage{subfigure}
\usepackage{multirow}
\usepackage{hyperref}
\usepackage{booktabs}
\usepackage{color}
\usepackage{textcomp}
\usepackage{caption}
\usepackage{wrapfig}
\usepackage[english]{babel}
\usepackage{blindtext}

\usepackage{booktabs}
\usepackage{makecell}
\usepackage[table]{xcolor}

\newcommand{\imgwidthe}{0.2\linewidth}

\renewcommand{\arraystretch}{1.2}

\begin{document}
%
%
%
%
\title{Domain and Geometry Agnostic CNNs for Left Atrium Segmentation in 3D Ultrasound}
\author{Markus A. Degel\inst{1,2}\and Nassir Navab\inst{1,3}\and Shadi Albarqouni\inst{1}}
\institute{Computer Aided Medical Procedures (CAMP), Technische Universit\"at M\"unchen, Munich, Germany
\and
TOMTEC Imaging Systems GmbH, Unterschleissheim, Germany
\and  
Whiting School of Engineering, Johns Hopkins University, Baltimore, USA
}

\maketitle

\begin{abstract}
	Segmentation of the left atrium and deriving its size can help to predict and detect various cardiovascular conditions. Automation of this process in 3D Ultrasound image data is desirable, since manual delineations are time-consuming, challenging and observer-dependent. Convolutional neural networks have made improvements in computer vision and in medical image analysis. They have successfully been applied to segmentation tasks and were extended to work on volumetric data. In this paper we introduce a combined deep-learning based approach on volumetric segmentation in Ultrasound acquisitions with incorporation of prior knowledge about left atrial shape and imaging device. The results show, that including a shape prior helps the domain adaptation and the accuracy of segmentation is further increased with adversarial learning.
\end{abstract}

\section{Introduction}
Quantification of cardiac chambers and their functions stay the most important objective of cardiac imaging \cite{2015_Lang}. Left atrium (LA) physiology and function have an impact on the whole heart performance and its size is a valuable indicator for various cardiovascular conditions, such as atrial fibrillation (AF), stroke and diastolic dysfunction \cite{2015_Lang}. Echocardiography, cardiac computed tomography (CCT) and cardiac magnetic resonance (CMR) are options to examine the heart. Echocardiography is the best choice, due to its wide availability, safety and good spatial and temporal resolution, without exposing the patients to harmful radiation. Volumetric measurements consider changes in all spatial dimensions, however, to obtain reproducible and accurate three-dimensional (3D) measurements, requires expert experience and is time consuming \cite{2017_vandenHoven}. Automated segmentation and quantification could help to reduce inter/intra-observer variabilities \cite{2016_Tsang} and might also save costs and time in echocardiographic laboratories \cite{2017_vandenHoven}.

\begin{figure}[t]
	\begin{center}
		\includegraphics[width=0.15\linewidth]{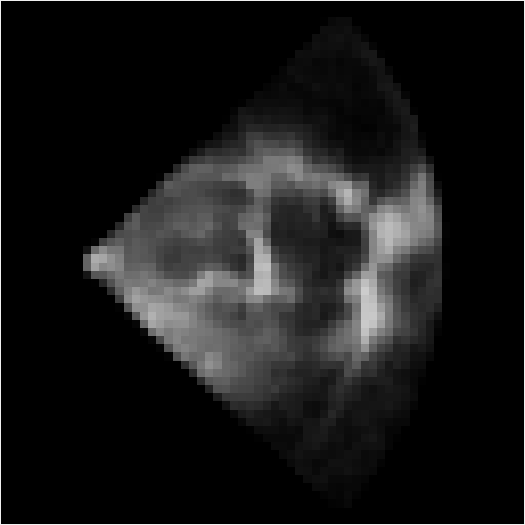}
		\includegraphics[width=0.15\linewidth]{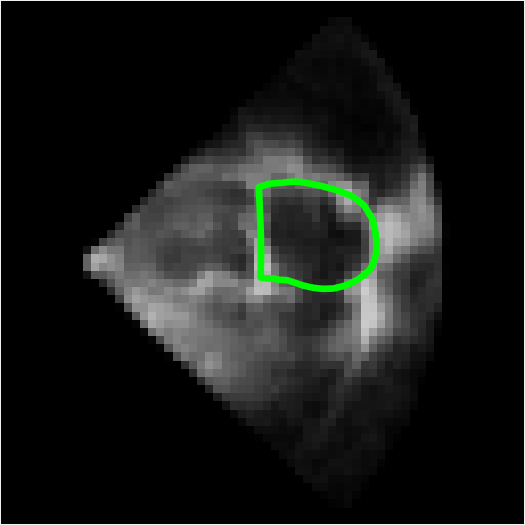}
		\includegraphics[width=0.15\linewidth]{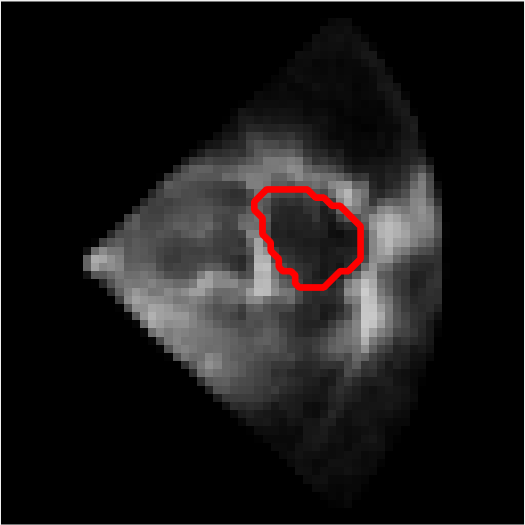}\\
		\includegraphics[width=0.15\linewidth]{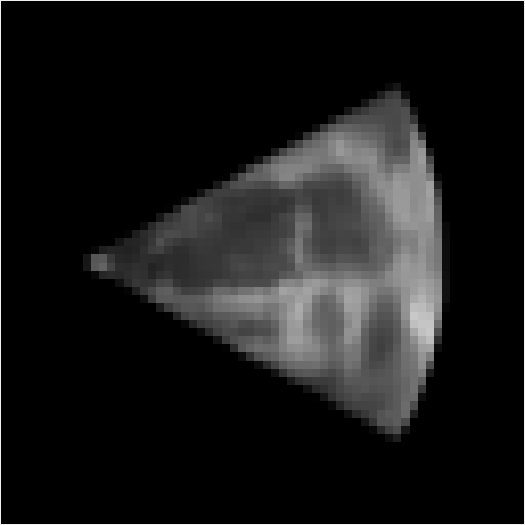}
		\includegraphics[width=0.15\linewidth]{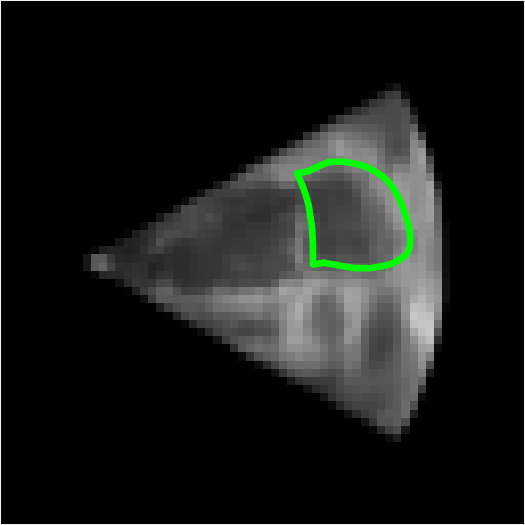}	\includegraphics[width=0.15\linewidth]{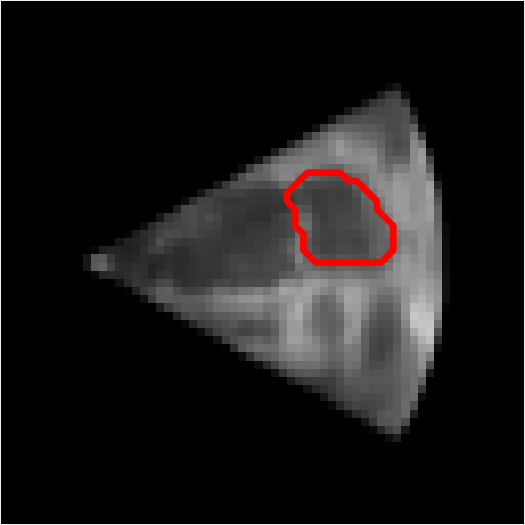}\\
		\includegraphics[width=0.15\linewidth]{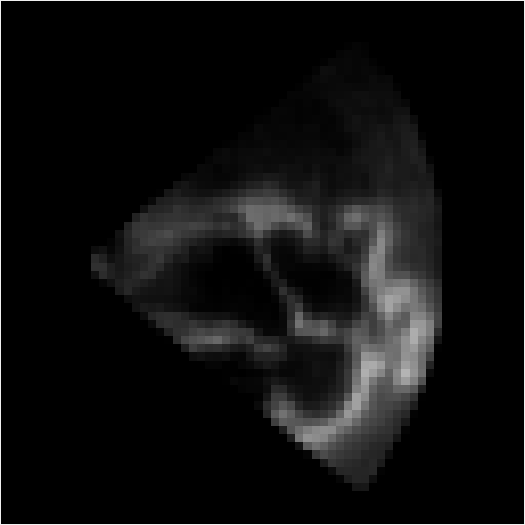}
		\includegraphics[width=0.15\linewidth]{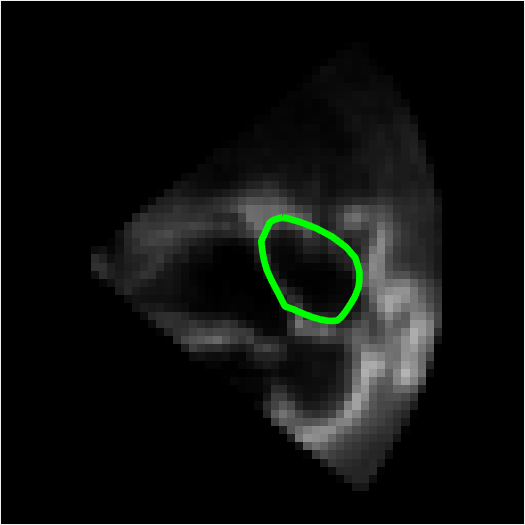}	\includegraphics[width=0.15\linewidth]{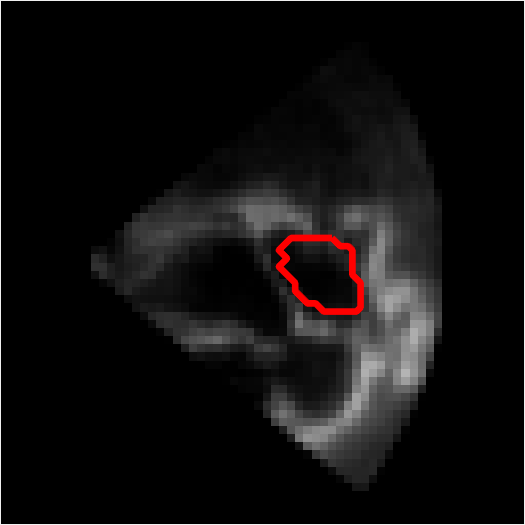}
		\caption{Row 1: device EPIQ 7C (dice coefficient: 0.74, training: Vivid E9), Row 2: device Vivid E9 (dice coefficient: 0.56, training: EPIQ 7C), Row 3: device iE33 (dice coefficient: 0.6, training: Vivid E9), Left: volume slice, Middle: ground truth delineation of LA, Right: prediction by C3 architecture (Table~\ref{table_results}).}
		\label{fig_data}
	\end{center}
\end{figure}

Previous automatic and semi-automatic approaches for LA segmentation have focused CCT and CMR as a planning and guidance tool for LA catheter interventions \cite{2016_Almeida}. For 3D Ultrasound (US), the left ventricle (LV) was the segmentation target, since its size and function remain the most important indication for a cardiac study \cite{2015_Knackstedt}. LA segmentation in 3D US data has not received much attention, apart from commercially available methods, which were also successfully validated against the gold standard CMR and CCT \cite{2011_Rohner,2013_Buechel}. Another approach exists, adapted from a segmentation framework for LV, based on B-spline explicit active surfaces \cite{2016_Almeida}. For transesophageal echocardiography (TEE), statistical shape models from a CT database were used \cite{2011_Voigt}. Those methods, however, require more or less manual interaction. Recently, fully automatic segmentation software for the left heart was validated against CMR \cite{2017_vandenHoven}.

Convolutional neural networks (CNN) and their special architectures of fully convolutional networks (FCN) have successfully been applied to the problem of medical image segmentation \cite{2015_Ronneberger}. Those networks are trained end-to-end, process the whole image and perform pixel-wise segmentation. The \textit{V-Net} extends this idea to volumetric MRI image data and enables 3D segmentation with the help of spatial convolutions, instead of processing the volumes slice-wise \cite{2016_Milletari}.

Automated segmentation in US images is challenging, due to artifacts (\emph{e.g} respiratory motion) or operator dependent errors (\emph{e.g} shadows, signal-dropouts). Including shape priors in this task can help algorithms to yield more accurate and anatomically plausible results. Oktay~\emph{et al.} introduced a way to incorporate such a prior with the help of an autoencoder network, that leads segmentation masks to follow an underlying shape representation \cite{2018_Oktay}.

Image data might be different (\emph{e.g} with respect to resolution, contrast), due to varying imaging protocols and device manufacturers \cite{2017_Kamnitsas}. Although the segmentation task is equivalent, neural networks perform poorly when applied to data that was not available during training. Generating ground truth maps and retraining a new model for each domain is not a scalable solution. The problem of models to generalize to new image data can be approached by domain adaptation. Kamnitsas~\emph{et al.} successfully introduced the application of unsupervised domain adaptation for brain lesion segmentation in different MRI databases, when an adversarial neural network was influencing the feature maps of a CNN, which was employed for the segmentation task \cite{2017_Kamnitsas}.

In this work, LA segmentation in 3D US volumes is performed with the help of neural networks. For the volumetric segmentation, \textit{V-Net} will be trained, combined with additional losses, taking into account the geometrical constrain introduced by the shape of the LA and the desired ability to generalize to different US devices and settings.

\section{Methodology}
\begin{figure}[t]
	\begin{center}
		\includegraphics[width=0.7\textwidth]{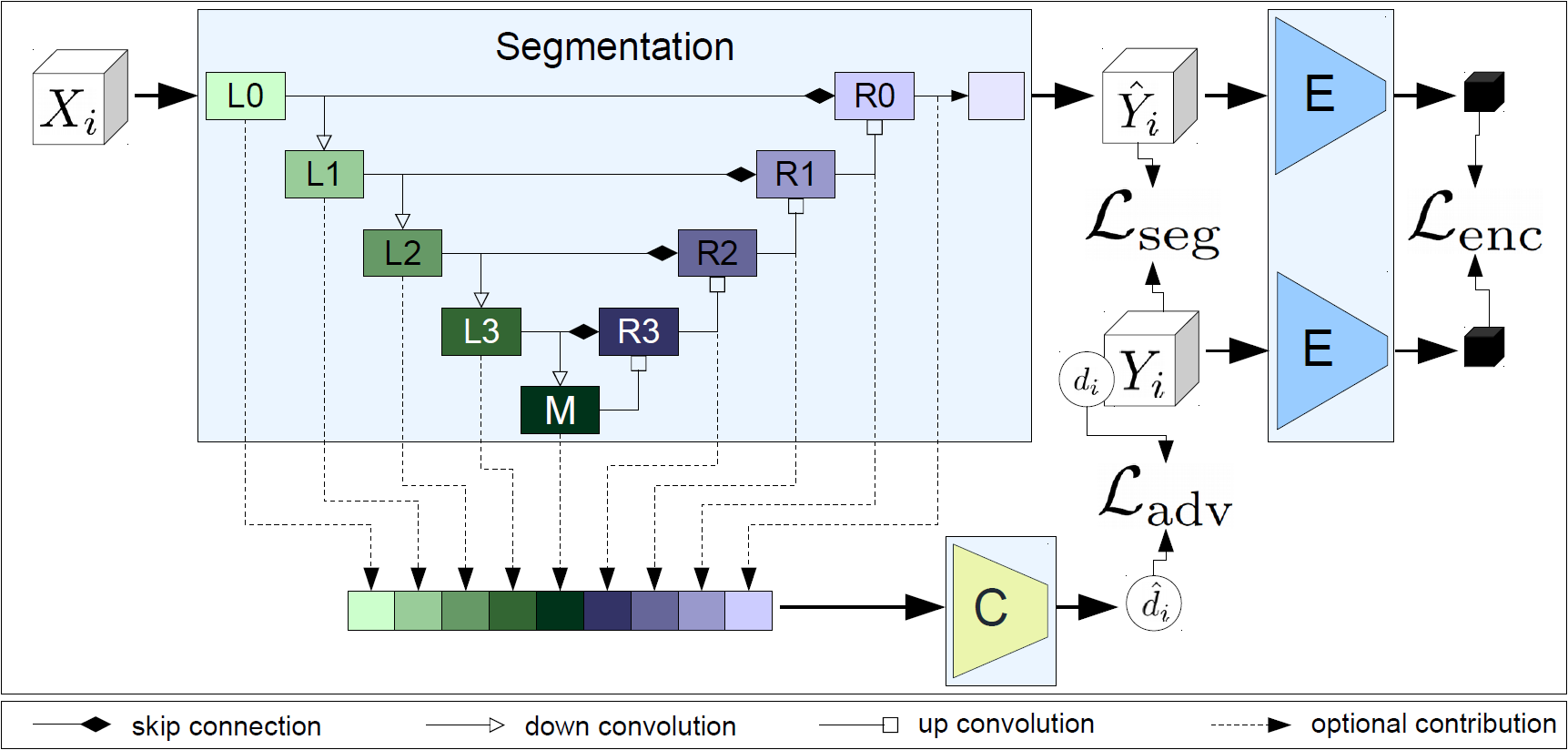}
		\caption{Overview of the combined architecture: Image data $X_i$ is processed by \textit{V-Net}\cite{2016_Milletari}. Dice-loss (Eq.~\ref{eq_diceloss}) is calculated from the resulting segmentation $\hat{Y}_i$ and the ground truth $Y_i$. Additionally, $\hat{Y}_i$ and $Y_i$ are encoded (E) to get the shape constrain. The feature maps of $X_i$ are extracted from V-Net to be processed in the classifier (C), which predicts a domain $\hat{d}_i$. Cross-entropy between $\hat{d}_i$ and the real domain $d_i$ determines the adversarial loss.}
		\label{fig_network}
	\end{center}
\end{figure}
Our framework, as depicted in Fig.~\ref{fig_network}, consists of three deep-learning blocks working jointly to achieve our objectives; LA segmentation with the help of incorporating a shape prior, and further being able to generalize well on different domains.

\paragraph{\textbf{Segmentation.}} For the segmentation task, we employ \textit{V-Net}\cite{2016_Milletari} as a fully convolutional network, which processes an image volume of size $n$, $X_i = \{x_1,...,x_n\}$, $x_i \in \mathcal{X}$ and yields a segmentation mask $\hat{Y}_i = \{\hat{y}_1,...,\hat{y}_n\}$, $\hat{y}_i \in \mathcal{\hat{Y}}$ in the original resolution. $\mathcal{X}$ represents the feature space of US acquisitions and $\mathcal{\hat{Y}}$ describes the probability of a voxel belonging to the segmentation.

The objective function of \textit{V-Net} is adapted to the segmentation task. It is based on the Dice coefficient (Eq.~\ref{eq_diceloss}), taking into account the possible imbalance of foreground to background, alleviating the need to re-weight samples.
\begin{equation}
\mathcal{L}\textsubscript{seg} = 1 - \frac{2 \cdot \sum_{i}^{}y_i \cdot \hat{y}_i}{\sum_{i}y_i^2 + \sum_{i}\hat{y}_i^2},
\label{eq_diceloss}
\end{equation}
with $\hat{y}_i$ being the prediction and $y_i$ the voxels of the ground truth $Y_i$ from the binary distribution $\mathcal{Y}$. 

\paragraph{\textbf{Shape prior.}} Incorporation of the shape prior to help the segmentation task is realized with the approach of \cite{2018_Oktay}. An autoencoder network is trained on the segmentation ground truth masks \textit{\textbf{Y}}. The encoder reduces the label to a latent, low resolution representation $E(Y_i)$ and the decoder tries to retrieve the original volume $Y_i$. Due to the resolution reduction of the encoder, the shape information is encoded in a compact fashion.

During training, the output of the segmentation network $\hat{Y}_i$ is passed to the encoder, along with the ground truth label $Y_i$. Based on a distance metric \textit{d}($\cdot$,$\cdot$), a loss between the latent codes of both inputs is calculated as
\begin{equation}
\mathcal{L}\textsubscript{enc} = d(E(Y_i),E(\hat{Y}_i)).
\end{equation}
The gradient is then back-propagated to the segmentation network.

\paragraph{\textbf{Domain adaptation.}} When a network is trained on one type of data $\mathcal{X}_S$ (source domain) and evaluated on another $\mathcal{X}_T$ (target domain), the performance is poor in most cases. Domain invariant features are desired to make the segmentation network perform well on different data sets. Kamnitsas~\emph{et al.} propose an approach to generate domain invariant features to increase the generalization capability \cite{2017_Kamnitsas}.

Processing an image volume in a CNN yields a latent representation $h_l(X_i)$ after convolutional layer $l$. If the network is not domain invariant, those feature maps contain, as expected, information about the data type (source or target domain). The idea to solve this issue is to train a classifier $C$, which takes feature maps of the segmentation network as input and returns a decision, if the input data was from source ($X_S$) or target ($X_T$) domain: $C(h_l(X_i)) = \hat{d}_i \in \{S,T\}$. The accuracy of this classifier with respect to the real domain $d_i$ is an indicator of how domain invariant the features are.

\paragraph{\textbf{Combination.}} The ideas introduced in the previous sections are now combined to exploit the advantages of the individual approaches (Fig.~\ref{fig_network}). The loss of the domain classifier is used as an adversarial loss term, since the goal of the segmentation network is to lower the classification accuracy (\emph{i.e} maximize the classification loss). The inability of the classifier to tell, which type of data was segmented means that the feature maps do not hold domain specific information. At the same time, the segmentation loss and the loss with respect to $E(Y_i)$ and $E(\hat{Y}_i)$ should be minimized.
This yields the following combined loss function:
\begin{equation}
\mathcal{L} = \mathcal{L}\textsubscript{seg} + \lambda\textsubscript{enc} \cdot \mathcal{L}\textsubscript{enc} - \lambda\textsubscript{adv} \cdot \mathcal{L}\textsubscript{adv}
\label{eq_combinedloss}
\end{equation}
with $\mathcal{L}\textsubscript{adv}$ being the binary cross entropy loss of the classifier $C$ and $\mathcal{L}\textsubscript{seg}$ describing the dice loss (Eq.~\ref{eq_diceloss}) with a weight regularization.

\section{Experiments and Results}
To evaluate the influence of the different loss terms, we apply it to 3D Ultrasound data to perform end-systolic LA segmentation. The network is trained with images and labels from one device and tested on different devices. 

\paragraph{\textbf{Dataset.}} The data available for this work are 3D transthoracic echocardiography (TTE) examinations taken from clinical routine. Multiple international centers contributed to a pool of 161 datasets, containing the LA ground truth segmentation in the recorded heart cycle, with the relevant phases for LA functionality (end-diastole, end-systole and pre-atrial contraction) identified.
\begin{table}[t]
	\begingroup
	\renewcommand{\arraystretch}{1.1}
	\centering
	\caption{Data device and set distribution. iE33 datasets are only used for evaluation. Resolutions are equidistant. For resolution and opening angles (azimuth \& elevation) the mean $\pm$ standard deviation in the respective set are shown.}
	\begin{tabular*}{\textwidth}{@{}l *{3}{@{\extracolsep{\fill}}c}}
		\toprule
		Property & EPIQ 7C & Vivid E9 & iE33 \\
		\midrule
		train/val/test & 33/7/27 & 39/8/32 & 0/0/15 \\
		Resolution (mm/voxel) & $0.95\pm0.10$ & $0.95\pm0.10$ & $0.96\pm0.11$ \\
		Azimuth (deg) & $87.1\pm4.7$ & $47.3\pm10.4$ & $80.2\pm0.0$ \\
		Elevation (deg) & $78.2\pm0.1$ & $47.4\pm10.5$ & $91.6\pm0.0$ \\
		\bottomrule
	\end{tabular*}
	\label{table_data}
	\endgroup
\end{table}

The image volumes were acquired with Ultrasound systems from GE (Vivid E9, GE Vingmed Ultrasound) and Philips (EPIQ 7C and iE33, Philips Medical Systems), each equipped with a matrix array transducer. Table \ref{table_data} shows the data distribution with respective splits into training, validation and testing sets. Since there are only 15 datasets for device iE33, those examinations are not used for training, but only for evaluation. For the data to be fed into the network, the size is down-sampled to 64 cubic volumes, preserving angles and ratios, by zero padding datasets as visible in Fig.~\ref{fig_data}.

\paragraph{\textbf{Implementation.}} Network architectures are implemented using the TensorFlow\footnote{\url{https://www.tensorflow.org/}} library (version 1.4) with GPU support. For our approach, the \textit{V-Net} architecture is adapted such that volumes of size 64x64x64 can be processed. The autoencoder network architecture is inspired from the one proposed in \cite{2018_Oktay}. To generate the input for the classifier, feature maps of V-Net at different levels have to be concatenated. The size of the bottleneck (4x4x4) for extracted feature maps is obtained by the (repeated) application of convolutions of filter size 2 and stride 2.
\begin{table}[t]
	\begingroup
	\renewcommand{\arraystretch}{1.1}
	\centering
	\caption{Training procedure details. Each training uses a learning rate decay of 0.99 after each epoch and a batch size of 4. $\textit{\textbf{X}} = \textit{\textbf{X}}_S \cup \textit{\textbf{X}}_T$, $\textit{\textbf{d}}$: domain labels.}
	\begin{tabular*}{\textwidth}{@{}l *{7}{@{\extracolsep{\fill}}c}}
		\toprule
		\# & Name (parameters) & optimizer & \makecell{learning \\ rate} & \makecell{weight \\ reg.} & epochs & data & label \\
		\midrule
		1 & Autoencoder ($\theta\textsubscript{ae}$) & Momentum $\beta$:0.9 & $5 \cdot 10^{-4}$ & 0.1 & 100 & $\textit{\textbf{Y}}_S$ & $\textit{\textbf{Y}}_S$ \\
		2 & Segmentation ($\theta\textsubscript{seg}$) & \makecell{Adam \\ $\beta_1$: 0.99, $\beta_2$: 0.999} & $1 \cdot 10^{-5}$ & $5 \cdot 10^{-4}$ & 50 & $\textit{\textbf{X}}_S$ & $\textit{\textbf{Y}}_S$ \\
		3 & Classifier ($\theta\textsubscript{adv}$) & SGD & $5 \cdot 10^{-5}$ & $1 \cdot 10^{-5}$ & 15 & $\textit{\textbf{X}}$ & $\textit{\textbf{d}}$ \\
		\multirow{2}{*}{4} & Combination\ref{eq_combinedloss} ($\theta\textsubscript{seg}$) & Momentum $\beta$:0.99 & $1 \cdot 10^{-5}$ & $5 \cdot 10^{-4}$ & \multirow{2}{*}{100} & $\textit{\textbf{X}}_S$ & $\textit{\textbf{Y}}_S$,$\textit{\textbf{d}}$ \\
		& Classifier ($\theta\textsubscript{adv}$) & SGD & $5 \cdot 10^{-5}$ & $1 \cdot 10^{-5}$ &  & $\textit{\textbf{X}}$ & $\textit{\textbf{d}}$ \\
		\bottomrule
	\end{tabular*}
	\label{table_training}
	\endgroup
\end{table}

\paragraph{\textbf{Training Details.}} 
The autoencoder network is trained before the combined training procedure, to obtain a meaningful latent representation for the shape prior. In the following training stages, the parameters of this network are frozen. 

The segmentation network is shortly pre-trained, as well as the classifier to introduce stability in the combined training. This way, the parameters of the single networks are pre-adjusted and training can focus on realizing the scenario defined by the settings of $\lambda\textsubscript{enc}$ and $\lambda\textsubscript{adv}$. Feature maps L0, L2, M, R2 and R0 of the segmentation network are extracted for the classifier (compare Fig.~\ref{fig_network}).

The combined training procedure starts by adding the loss term for incorporation of the shape prior to the segmentation loss. Adversarial influence begins after $e\textsubscript{adv}$ = 10 epochs of combined training. $\lambda\textsubscript{adv}$ is increased linearly until its maximum $\lambda\textsubscript{adv,max}$ = 0.001 ($\lambda\textsubscript{adv}~=~\min((e~-~e\textsubscript{adv}~+~1)~\cdot~\alpha,~1)~\cdot~\lambda\textsubscript{adv,max}$, with $e$ the current epoch and $\alpha$~=~0.1 the adversarial influence growth factor). While the combined training adjusts the parameters of the segmentation network $\theta\textsubscript{seg}$ only, the classifier parameters $\theta\textsubscript{adv}$ are continued to be trained in parallel to retain a potent adversarial loss term. A training overview is given in Table~\ref{table_training}. Different parameters for the combined network are recorded in Table~\ref{table_results}.

\paragraph{\textbf{Evaluation.}} The segmentation network returns a volume $\hat{Y}_i$ of probabilities for the voxels to belong to the foreground, \emph{i.e} the segmentation of the LA. The threshold for the cutoff probability to obtain a binary segmentation mask is determined by the best Dice coefficient on the validation set, from which the biggest connected component is selected as the final LA segmentation.
\begin{table}[t]
	\begingroup
	\renewcommand{\arraystretch}{1.0}
	\centering
	\caption{Results for ES LA segmentation. For completeness, results of \textit{ACNN} and \textit{V-Net} are reported. C1: $\lambda\textsubscript{adv} = 0$, $d$: L2-distance. C2: $\lambda\textsubscript{adv} = 0$, $d$: ACD. C3: $\lambda\textsubscript{adv} = 0.001$, $d$: ACD. C1,C2 \& C3: $\lambda\textsubscript{enc} = 0.001$. Format: mean~$\pm$~std.}
	\begin{tabular*}{\textwidth}{@{}l *{6}{@{\extracolsep{\fill}}c}@{}}
		\toprule
		Training & Test & \textit{V-Net}\cite{2016_Milletari} & \textit{ACNN}\cite{2018_Oktay} & C1 & C2 & C3 \\
		\midrule
		\multicolumn{7}{l}{\textbf{Mean Surface Distance (MSD)}} \\
		EPIQ 7C & \makecell[r]{EPIQ 7C \\ Vivid E9 \\ iE33} &
		\makecell{\textbf{1.16$\pm$0.88} \\ $3.56\pm1.71$ \\ $1.44\pm0.77$} &
		\makecell{$1.35\pm1.19$ \\ $10.67\pm7.29$ \\ \textbf{1.38$\pm$0.40}} &
		\makecell{$1.26\pm0.69$ \\ $3.87\pm3.06$ \\ $2.33\pm2.38$} &
		\makecell{$1.27\pm0.69$ \\ $2.42\pm1.32$ \\ $1.94\pm1.49$} &
		\makecell{$1.21\pm0.60$ \\ \textbf{2.01$\pm$1.63} \\ $1.44\pm0.35$} \\
		\arrayrulecolor{black!=10}\specialrule{1pt}{0pt}{0.5pt}
		Vivid E9 & \makecell[r]{EPIQ 7C \\ Vivid E9 \\ iE33} &
		\makecell{$2.87\pm1.53$ \\ \textbf{0.94$\pm$0.59} \\ $4.72\pm4.86$} &
		\makecell{$4.39\pm1.33$ \\ $1.57\pm0.87$ \\ $3.28\pm2.22$} &
		\makecell{$2.12\pm0.96$ \\ $1.18\pm0.38$ \\ $4.18\pm3.36$} &
		\makecell{$1.87\pm0.96$ \\ $1.12\pm0.37$ \\ $3.18\pm2.88$} &
		\makecell{\textbf{1.59$\pm$1.04} \\ $1.18\pm0.37$ \\ \textbf{2.62$\pm$1.46}} \\
		\arrayrulecolor{black!=30}\specialrule{1pt}{0pt}{0.5pt}
		\multicolumn{7}{l}{\textbf{Hausdorff Distance (HD)}} \\
		EPIQ 7C & \makecell[r]{EPIQ 7C \\ Vivid E9 \\ iE33} &
		\makecell{\textbf{4.46$\pm$2.73} \\ $7.66\pm2.94$ \\ \textbf{4.06$\pm$1.21}} &
		\makecell{$5.52\pm3.15$ \\ $16.87\pm8.92$ \\ $5.03\pm1.39$} &
		\makecell{$5.51\pm2.31$ \\ $8.21\pm5.06$ \\ $5.60\pm2.86$} &
		\makecell{$5.33\pm2.07$ \\ $5.79\pm2.21$ \\ $4.98\pm2.02$} &
		\makecell{$4.92\pm1.60$ \\ \textbf{5.46$\pm$3.36} \\ $4.70\pm0.91$} \\
		\arrayrulecolor{black!=10}\specialrule{1pt}{0pt}{0.5pt}
		Vivid E9 & \makecell[r]{EPIQ 7C \\ Vivid E9 \\ iE33} &
		\makecell{$10.82\pm3.80$ \\ \textbf{3.67$\pm$2.29} \\ $9.52\pm6.44$} &
		\makecell{$13.63\pm2.87$ \\ $7.09\pm3.21$ \\ $11.60\pm3.72$} &
		\makecell{$8.09\pm2.88$ \\ $5.41\pm1.84$ \\ $9.08\pm3.64$} &
		\makecell{$7.31\pm2.51$ \\ $5.05\pm1.70$ \\ $7.13\pm3.49$} &
		\makecell{\textbf{5.47$\pm$2.45} \\ $5.14\pm1.26$ \\ \textbf{6.63$\pm$2.25}} \\
		\arrayrulecolor{black!=30}\specialrule{1pt}{0pt}{0.5pt}
		\multicolumn{7}{l}{\textbf{Dice Coefficient (DC)}} \\
		EPIQ 7C & \makecell[r]{EPIQ 7C \\ Vivid E9 \\ iE33} &
		\makecell{\textbf{0.75$\pm$0.17} \\ $0.10\pm0.21$ \\ $0.57\pm0.31$} &
		\makecell{$0.69\pm0.20$ \\ $0.15\pm0.25$ \\ $0.64\pm0.11$} &
		\makecell{$0.74\pm0.10$ \\ $0.33\pm0.27$ \\ $0.55\pm0.19$} &
		\makecell{$0.73\pm0.11$ \\ $0.32\pm0.26$ \\ $0.59\pm0.19$} &
		\makecell{$0.74\pm0.10$ \\ \textbf{0.55$\pm$0.23} \\ \textbf{0.67$\pm$0.08}} \\
		\arrayrulecolor{black!=10}\specialrule{1pt}{0pt}{0.5pt}
		Vivid E9 & \makecell[r]{EPIQ 7C \\ Vivid E9 \\ iE33} &
		\makecell{$0.56\pm0.15$ \\ \textbf{0.80$\pm$0.08} \\ $0.49\pm0.37$} &
		\makecell{$0.32\pm0.18$ \\ $0.69\pm0.11$ \\ \textbf{0.50$\pm$0.16}} &
		\makecell{$0.59\pm0.14$ \\ $0.73\pm0.07$ \\ $0.38\pm0.25$} &
		\makecell{$0.62\pm0.17$ \\ $0.74\pm0.08$ \\ $0.46\pm0.27$} &
		\makecell{\textbf{0.63$\pm$0.17} \\ $0.73\pm0.09$ \\ $0.46\pm0.19$} \\
		\arrayrulecolor{black!=10}\specialrule{1pt}{0pt}{0pt}
	\end{tabular*}
	\label{table_results}
	\endgroup
\end{table}

Segmentation metrics \cite{2016_Almeida,2018_Oktay} are reported in Table~\ref{table_results} for the recommended phase of LA segmentation (end-systole ES \cite{2015_Lang}). C1 describes the \textit{V-Net} architecture with the additional loss term $\mathcal{L}\textsubscript{enc}$, calculated from the L2-distance ($d(p,q) = \Vert p - q\Vert^2_2$). To investigate the influence of a different distance metric, C2 uses the angular cosine distance (ACD, $d(p,q) = 1 - \frac{\sum_{i}p_i \cdot q_i}{\Vert p\Vert_2 \cdot \Vert q\Vert_2}$). C3 leverages the better performing distance metric (ACD) with the adversarial loss $\mathcal{L}\textsubscript{adv}$. We define statistical significance based on the paired two-sample t-test on a 5~\% significance level.

When training on EPIQ 7C, \textit{V-Net} performs better than the other architectures on the same device. However, those margins are not statistically significant (MSD: $p = 0.65$, HD: $p = 0.24$, DC: $P = 0.66$), compared to C3. The increased performance of C3 compared to \textit{V-Net} and \textit{ACNN} is significant with respect to all metrics.

Vivid E9 training yields \textit{V-Net} with the best performance on the same device, with statistical significance on all metrics. C3 is significantly outperforming \textit{V-Net} on EPIQ 7C in terms of MSD and HD.

No significant differences are observable on the evaluation of device iE33. Independent of the distance metric utilized, an improvement in generalizability is observable compared to \textit{V-Net} when the shape prior is included (C1 \& C2).

\section{Discussion and Conclusion}
The results show, that including a shape prior for the segmentation task is actually helpful for domain adaptation. The adversarial loss, represented by the classifier accuracy is further improving the ability of the network to generalize. We show that the combination of loss terms for different objectives can have a great potential for domain adaptation. The type of distance metric utilized for the geometrical constrain would be an interesting subject to further investigate. An average dice coefficient improvement to \textit{V-Net} of 8.5~\% was achieved on our objective of LA segmentation.

\section*{Acknowledgment}
We would like to thank Ozan Oktay for sharing his implementation of the \textit{ACNN} architecture as a baseline of our project. Further, we thank Georg Schummers and Matthias Friedrichs from TOMTEC Imaging Systems GmbH for their support and helpful discussions.

\bibliography{LA_Segmentation,biblio-macros}
\bibliographystyle{splncs03}
\newpage

\section*{Appendix}
This section contains some supplementary images to show the different recorded heart cycle phases, the down-sampling process and segmentation results.
\vspace*{2\baselineskip}

\noindent
\textbf{Ultrasound Data}
\begin{figure}[h]
	\begin{center}
		\includegraphics[width=0.32\linewidth]{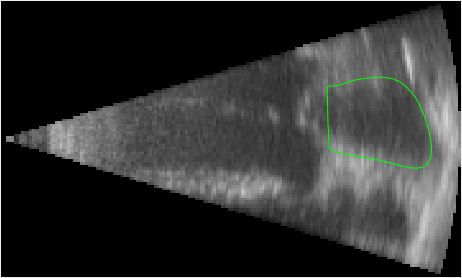}
		\includegraphics[width=0.32\linewidth]{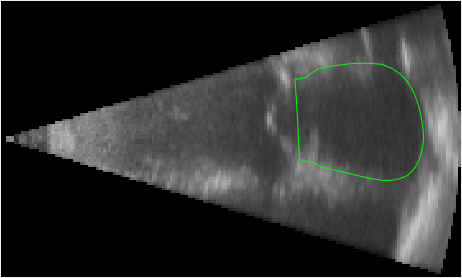}
		\includegraphics[width=0.32\linewidth]{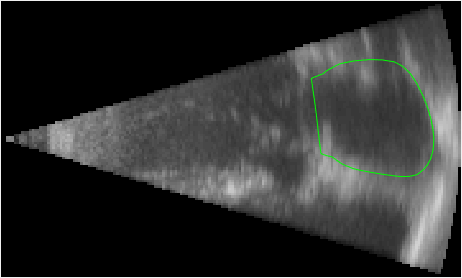}\\
		\vspace*{0.1\baselineskip}
		\includegraphics[width=0.32\linewidth]{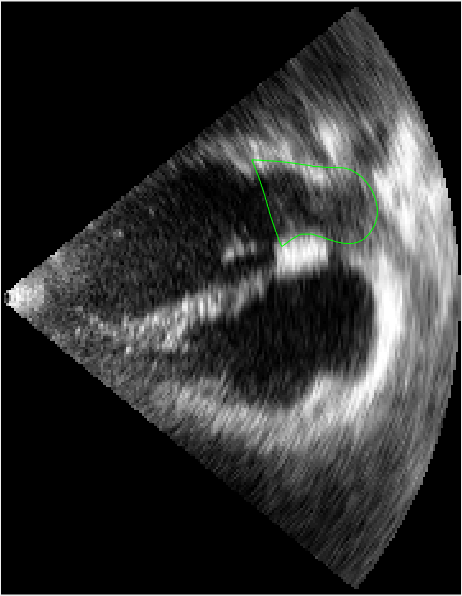}
		\includegraphics[width=0.32\linewidth]{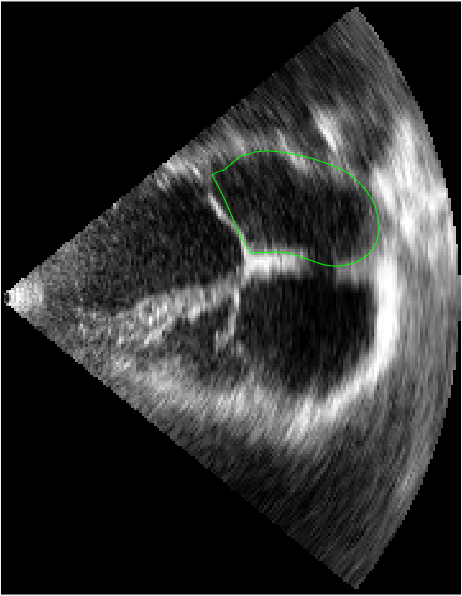}
		\includegraphics[width=0.32\linewidth]{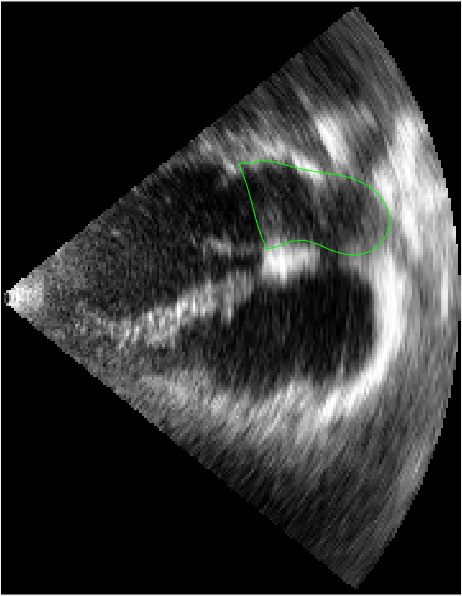}
		\caption{Top: device Vivid E9, Bottom: device EPIQ 7C, Left: end-diastole, Middle: end-systole, Right: pre-atrial contraction.}
	\end{center}
\end{figure}
\newpage

\noindent
\textbf{Preprocessing: down-sampling}
\begin{figure}[h]
\begin{center}
	\includegraphics[width=0.32\linewidth]{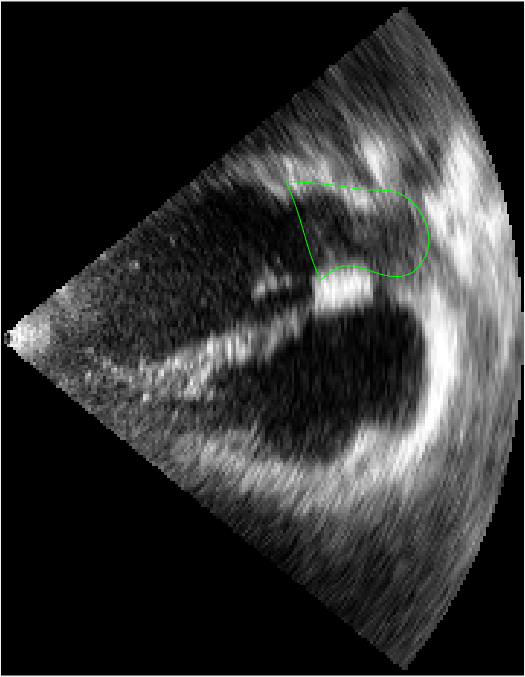}
	\includegraphics[width=0.32\linewidth]{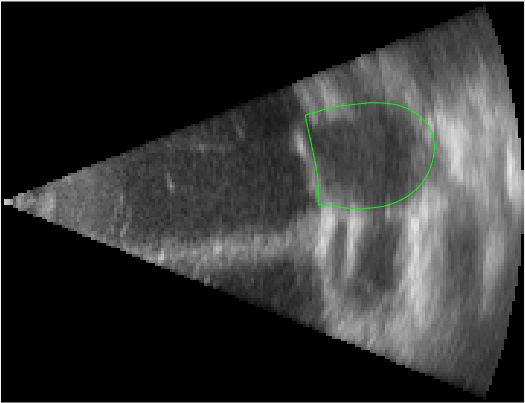}
	\includegraphics[width=0.32\linewidth]{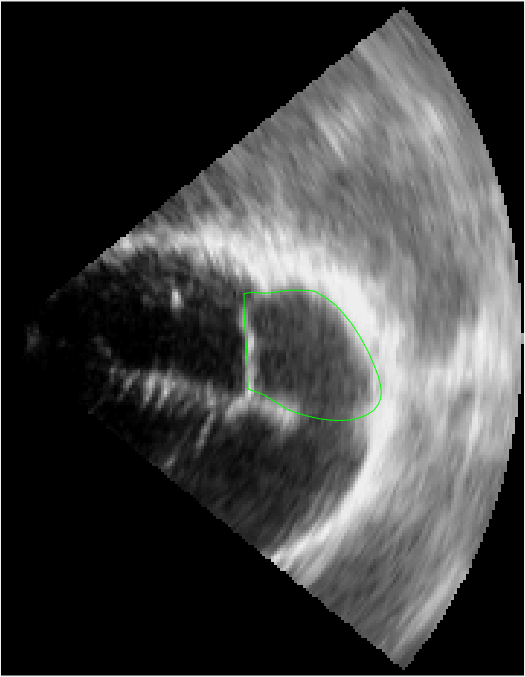}\\
	\vspace*{0.1\baselineskip}
	\includegraphics[width=0.32\linewidth]{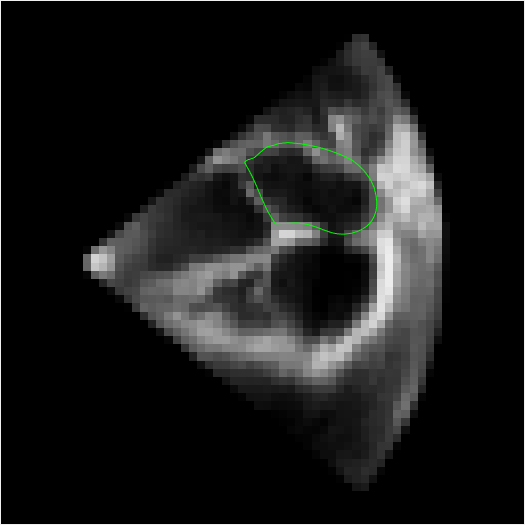}
	\includegraphics[width=0.32\linewidth]{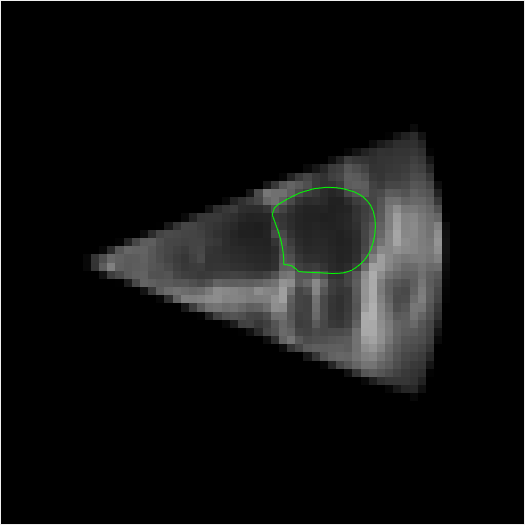}
	\includegraphics[width=0.32\linewidth]{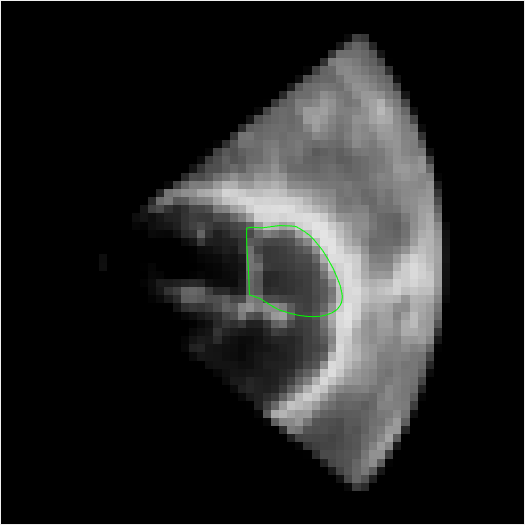}
	\caption{Top: high resolution, Bottom: down-sampled to 64x64x64 with zero padding, Left: device EPIQ 7C, Middle: device Vivid E9, Right: device iE33.}
\end{center}
\end{figure}
\newpage

\noindent
\textbf{Segmentation Result Images}
\begin{figure}[h]
	\begin{center}
		\includegraphics[width=\imgwidthe]{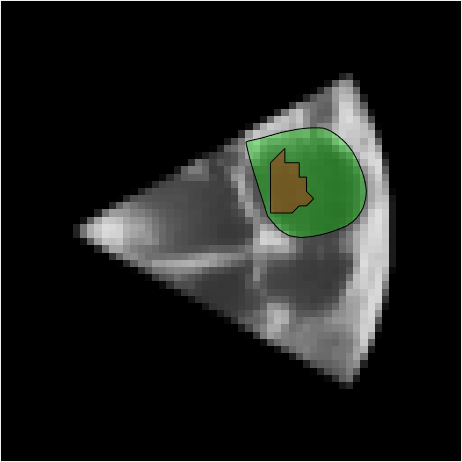}
		\includegraphics[width=\imgwidthe]{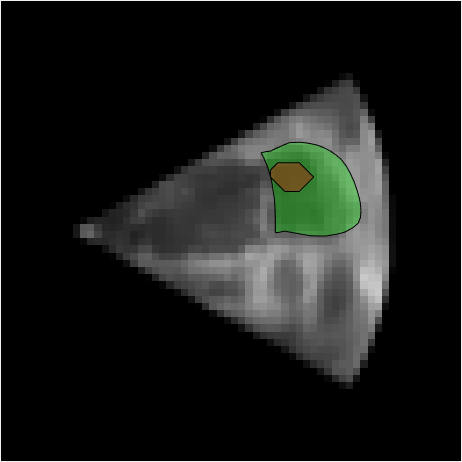}
		\includegraphics[width=\imgwidthe]{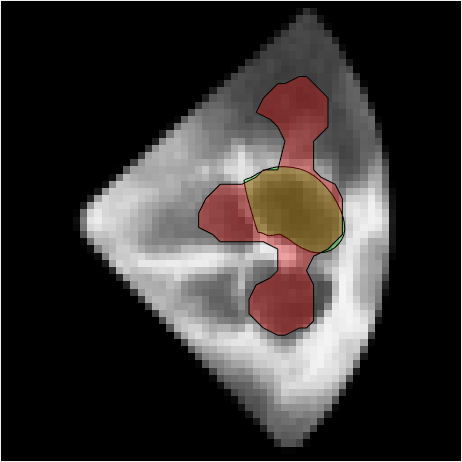}
		\includegraphics[width=\imgwidthe]{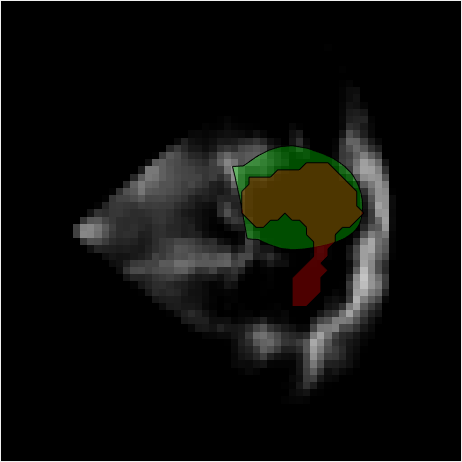}\\
		\vspace*{0.1\baselineskip}
		\includegraphics[width=\imgwidthe]{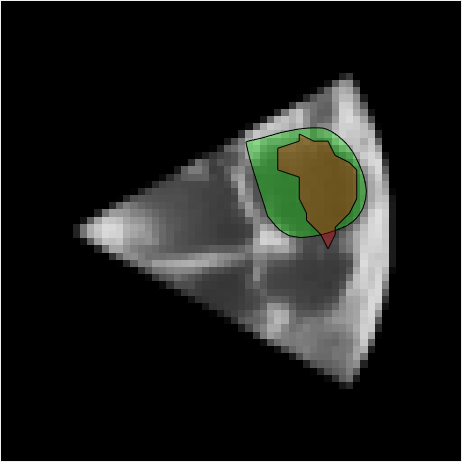}
		\includegraphics[width=\imgwidthe]{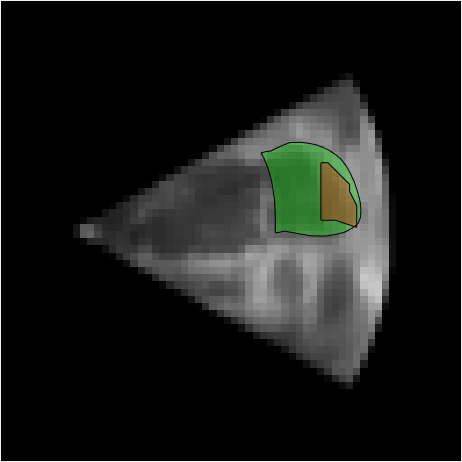}
		\includegraphics[width=\imgwidthe]{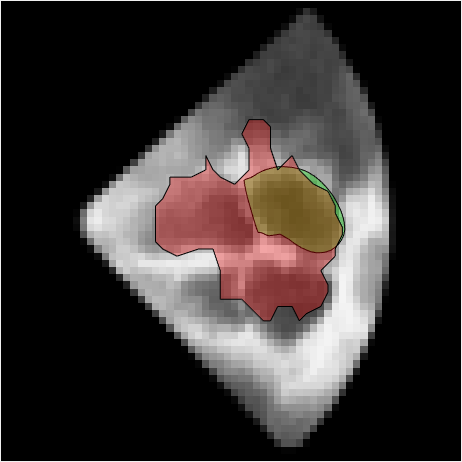}
		\includegraphics[width=\imgwidthe]{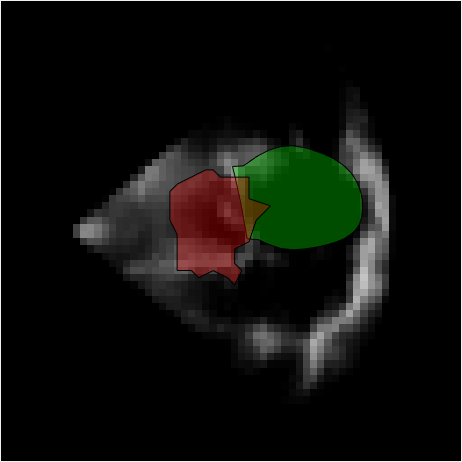}\\
		\vspace*{0.1\baselineskip}
		\includegraphics[width=\imgwidthe]{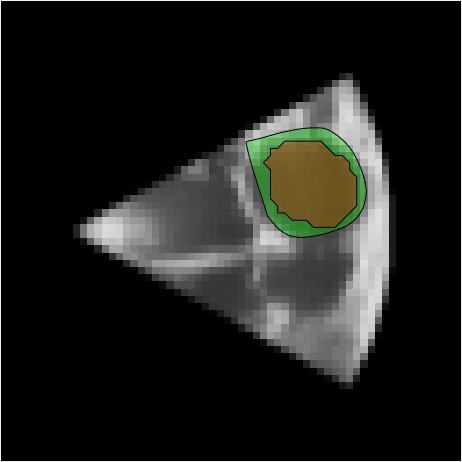}
		\includegraphics[width=\imgwidthe]{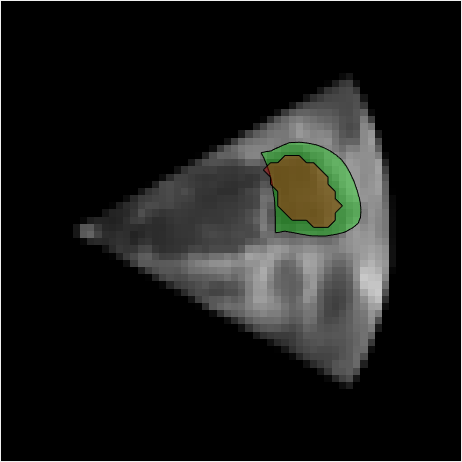}
		\includegraphics[width=\imgwidthe]{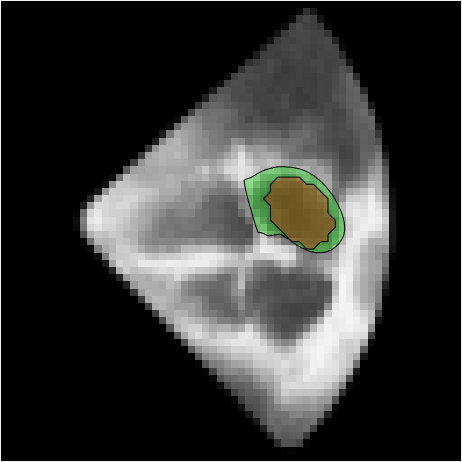}
		\includegraphics[width=\imgwidthe]{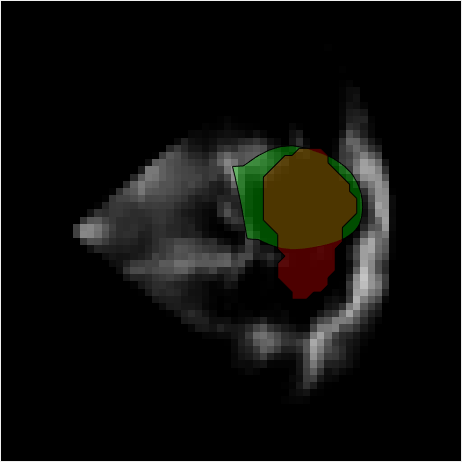}\\
		\vspace*{0.1\baselineskip}
		\includegraphics[width=\imgwidthe]{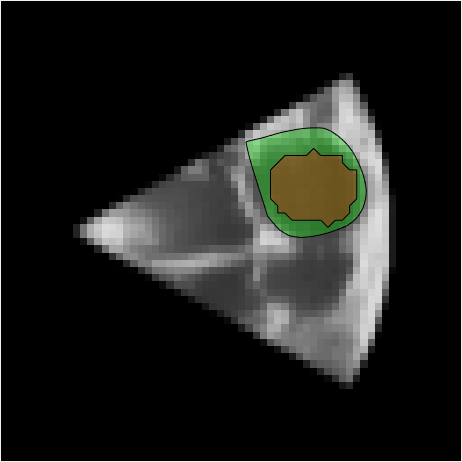}
		\includegraphics[width=\imgwidthe]{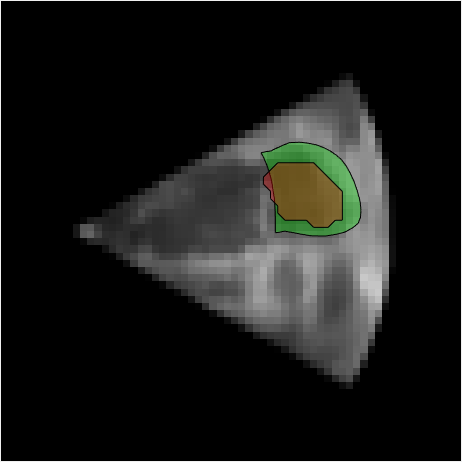}
		\includegraphics[width=\imgwidthe]{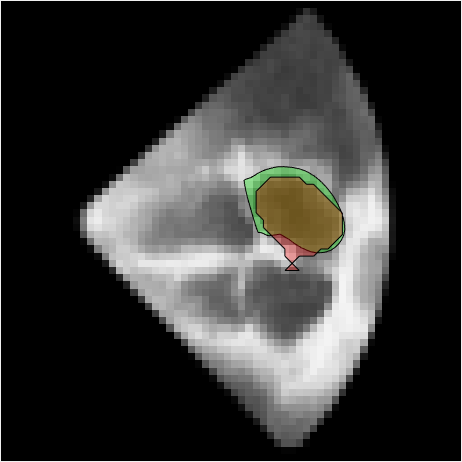}
		\includegraphics[width=\imgwidthe]{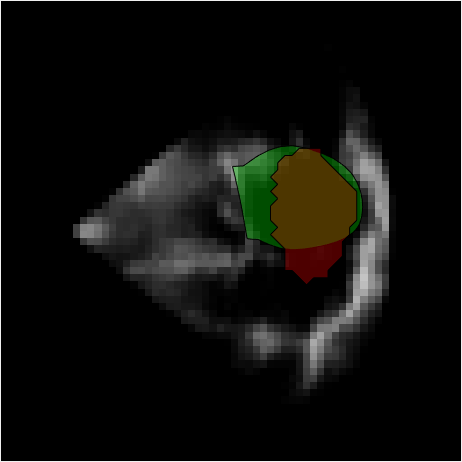}\\
		\vspace*{0.1\baselineskip}
		\includegraphics[width=\imgwidthe]{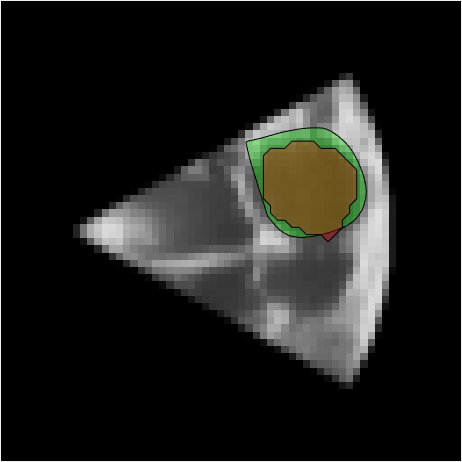}
		\includegraphics[width=\imgwidthe]{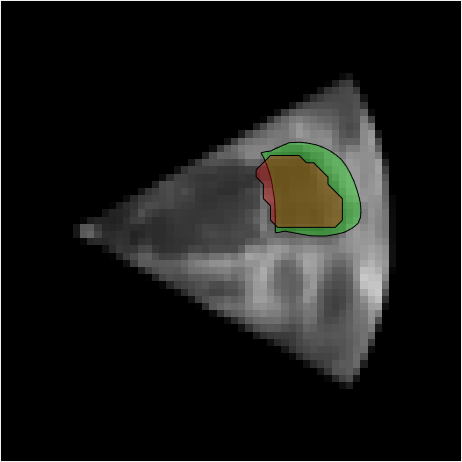}
		\includegraphics[width=\imgwidthe]{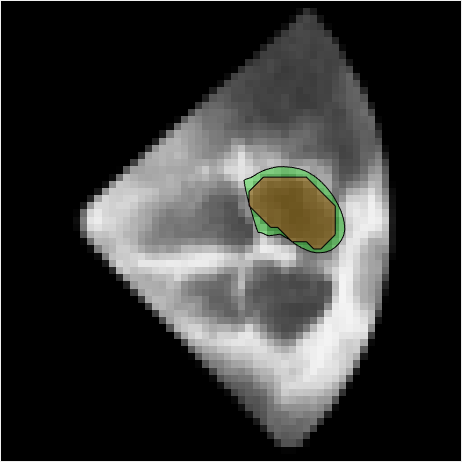}
		\includegraphics[width=\imgwidthe]{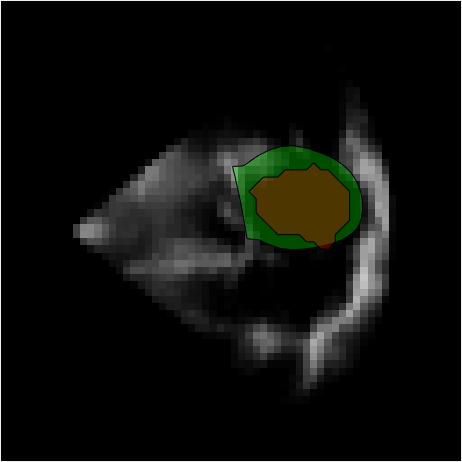}
		\caption{Row~1:~\textit{V-Net}, Row~2:~\textit{ACNN}, Row~3:~C1, Row~4:~C2, Row~5:~C3, Column~1~\&~2:~test result device Vivid E9 (training device EPIQ 7C), Column~3~\&~4:~test result device EPIQ 7C (training device Vivid E9).}
	\end{center}
\end{figure}
\newpage

\noindent
\textbf{Result plots}
\begin{figure}[h]
	\begin{center}
		\includegraphics[width=0.49\linewidth]{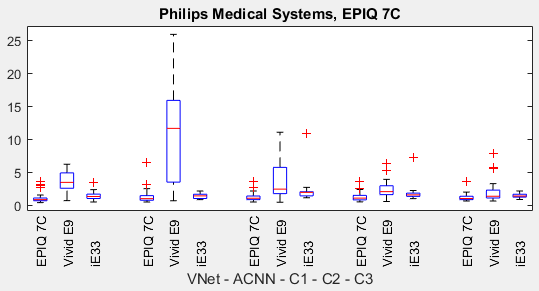}
		\includegraphics[width=0.49\linewidth]{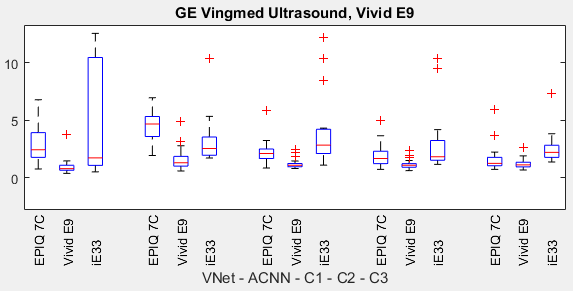}\\
		\vspace*{0.1\baselineskip}
		\includegraphics[width=0.49\linewidth]{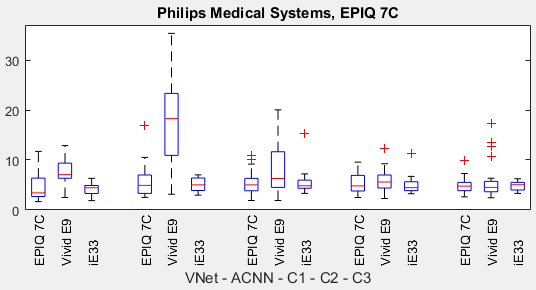}
		\includegraphics[width=0.49\linewidth]{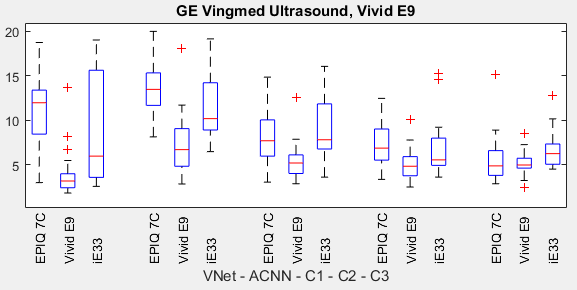}\\
		\vspace*{0.1\baselineskip}
		\includegraphics[width=0.49\linewidth]{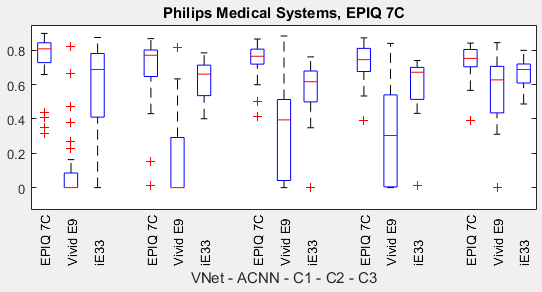}
		\includegraphics[width=0.49\linewidth]{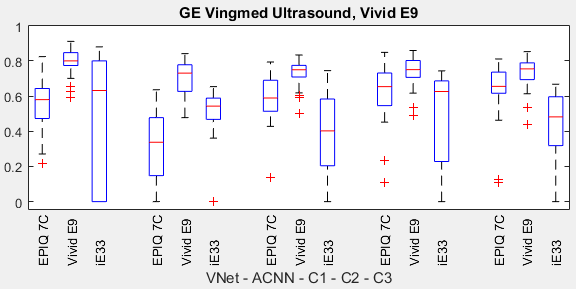}\\
		\caption{Top: Mean Surface Distance, Middle: Hausdorff Distance, Bottom: Dice Coefficient, Left: EPIQ 7C training, Right: Vivid E9 training.}
	\end{center}
\end{figure}

\end{document}